\documentclass{article}
\usepackage{graphicx} 
\usepackage{amsmath}

\date{}

\title{Generative AI and the Productivity Divide: Human–AI Complementarities in Education and Knowledge Work (Working Paper)}
\author{Bharat Anand, Leonard N. Stern School of Business, New York University \\ Lihi Idan, Industrial and Systems Engineering Department, Texas A\&M University}

\usepackage[margin=1in]{geometry}
\begin{document}

\maketitle

\begin{abstract}
     Generative Artificial Intelligence (GenAI) is transforming how firms create, process, and apply knowledge, yet little is known about the heterogeneity of its productivity effects across users. We report results from a randomized controlled experiment in which participants—analogs of early-career knowledge workers—were assigned to self-study a technical domain using either traditional resources or large-language-model (LLM) assistance. On average, GenAI access significantly increased task performance, but the distribution of gains was highly uneven. Improvements were not predicted by GPA or prior knowledge, but by \textit{AI Interaction Competence (AIC)} --- the ability to elicit, filter, and verify model outputs. High-AIC participants realized outsized gains; low-AIC participants saw limited or even negative marginal returns. A scaffolding intervention (conceptual maps) reduced outcome variance, indicating that standardized workflows can mitigate inequality in AI-mediated performance. We interpret these findings through the lens of human–AI complementarities: GenAI raises mean productivity while introducing a new axis of capability inequality. Managerially, firms should pair GenAI access with short AIC micro-training and simple standard operating procedures to capture value consistently and avoid uneven adoption outcomes.
\end{abstract}

\section{Introduction}

Generative AI (GenAI) is transforming how knowledge work is produced and managed.
Across industries, large language models (LLMs) are being integrated into research, analysis, communication, and decision-support workflows. Early field and laboratory studies document substantial productivity gains from GenAI adoption—but also striking variation in who benefits and by how much. Some workers experience dramatic improvements in speed and quality, while others see limited or even negative returns. Understanding the sources of this heterogeneity has become central to research on digital transformation, human capital, and the future of work.

A long tradition in management research shows that technology alone does not create value: its payoff depends on complementary human and organizational capabilities \cite{brynjolfsson2003computing,melville2004review,aral2012information,tambe2019ai}. GenAI heightens this complementarity. Unlike earlier forms of automation that standardized routine tasks, LLMs require users to engage in iterative problem solving—formulating prompts, interpreting probabilistic outputs, and verifying content quality. As a result, productivity depends not only on access to AI tools but also on a set of interactional skills that are tacit, unevenly distributed, and rarely taught.

We refer to this capability as AI Interaction Competence (AIC)—the ability to elicit, evaluate, and apply model outputs effectively. Conceptually, AIC is a new dimension of human capital that determines how individuals translate AI access into performance. Workers high in AIC can use GenAI to augment their expertise and extend their productivity frontier; those low in AIC may fail to extract value or risk compounding errors. This dynamic creates a new form of the productivity divide: the technology raises mean performance but may also widen dispersion in outcomes.

These dynamics are particularly salient in organizations that rely on knowledge-intensive work—consulting, finance, analytics, design, engineering, and professional services—where employees increasingly combine human judgment with AI-generated content. In such contexts, the performance impact of GenAI depends less on raw computational power and more on employees’ ability to collaborate productively with the technology. As firms adopt AI tools at scale, leaders face a dual challenge: capturing the productivity upside while preventing uneven capability development that could fragment teams or deepen performance inequality.

From a workforce development and training perspective, this challenge parallels historical shifts in organizational learning. Just as spreadsheet adoption required training in modeling and interpretation, GenAI adoption requires training in prompting, critical evaluation, and the structured use of outputs. Yet most firms lack systematic methods for diagnosing and developing these competencies. Understanding how individuals acquire AIC—and which interventions improve it—can inform the design of scalable training programs that accelerate learning and reduce variance in AI-enabled performance.

This study examines the micro-mechanisms behind that divide. Using a randomized controlled experiment that simulates early-career knowledge work—learning, synthesizing, and applying new technical information—we test how GenAI affects both average productivity and performance variance. We then investigate whether simple managerial interventions, such as structured guidance and process scaffolds, can reduce dispersion without lowering mean performance. 

Our contributions are threefold. First, we provide causal micro-evidence on the productivity effects of GenAI and show how individual capability heterogeneity drives uneven returns. Second, we introduce AI Interaction Competence as a theoretically grounded construct linking human capital and digital-complementarity literatures. Third, we demonstrate that process design—conceptual scaffolding and micro-training—can serve as a practical lever for managing variance in AI-mediated performance. Together, these insights reframe GenAI adoption as a problem of capability design: organizations must align technology, human skill, and workflow structure to capture value consistently.

From a managerial standpoint, the implications are direct. Firms seeking to harness GenAI should view access as the beginning, not the endpoint, of digital transformation. They can enhance value realization by embedding short micro-trainings to cultivate AIC, designing workflows that formalize how AI outputs are verified and applied, and using performance feedback to identify where human–AI interactions break down. In doing so, organizations can achieve two outcomes that are often treated as tradeoffs—higher average productivity and more equitable distribution of gains. GenAI’s promise lies not only in automation, but in reconfiguring how human capability and machine intelligence co-produce value.
\section{Literature Review}

Management research has long emphasized that technology creates value only when combined with complementary human and organizational capabilities \cite{brynjolfsson2003computing,melville2004review,aral2012information,tambe2019ai}.  Generative AI (GenAI) extends this insight.  Unlike earlier automation waves that standardized work, GenAI enables interactive augmentation—tools that rely on users’ ability to articulate objectives, interpret probabilistic outputs, and iteratively refine responses.  This shift heightens the importance of individual-level complementarities between digital tools and human skill \cite{brynjolfsson2017machine,rai2019misq}.  

Economic theories of \emph{skill-biased technical change} explain why such technologies often increase average productivity while widening inequality.  As automation substitutes for routine tasks, the relative value of abstract reasoning, social coordination, and adaptability rises \cite{acemoglu2020robots,acemoglu2018ai,autor2024middle,deming2017social}.  GenAI continues this trajectory by automating linguistic and cognitive subtasks but amplifying the advantage of individuals who can direct, evaluate, and verify model outputs.  We conceptualize this capability as \textit{AI Interaction Competence (AIC)}—a form of human capital that determines who captures the value of AI augmentation.

Emerging empirical evidence underscores these mechanisms.  Field experiments in customer service, writing, and consulting show that GenAI boosts mean productivity but introduces new variance across users \cite{brynjolfsson2023generative,eloundou2023gpts,dellacqua2025cybernetic}.  In each case, human capabilities and organizational scaffolds determine realized value.  Our study complements this literature by providing controlled, causal micro-evidence: it isolates how differences in AIC levels shape productivity outcomes and demonstrates that lightweight process interventions—conceptual scaffolding and training—reduce variance without lowering mean performance.

By integrating perspectives from information systems, economics, and organizational behavior, we position GenAI adoption as both a technological and managerial challenge: success depends on pairing AI capability with human competence and process structure.  This framing links our work to central debates in business analytics, operations management, and strategy on how organizations can capture the uneven gains of digital innovation.

\section{Research Questions}

This study is guided by four interrelated research questions designed to assess not only the average effectiveness of generative AI in organizations, but also the distributional consequences for task performance (and, by implication, productivity) and the mechanisms behind them. Together, these questions speak to a core management debate: does GenAI lift productivity uniformly or increase performance dispersion?

\textbf{RQ1: Do knowledge workers’ perceptions of their own knowledge align with their actual knowledge levels?}

A growing literature in management and behavioral science highlights the importance of calibration --- the ability of individuals to assess their own strengths and weaknesses. Poor calibration can undermine learning because participants may misallocate effort or overestimate their preparedness. We therefore examine whether knowledge workers’ self-assessments of task-relevant knowledge track their measured performance, and whether systematic biases exist. This serves both as a validity check for our measures and as a window into how knowledge workers approach AI-mediated training.

\textbf{RQ2: To what degree do generative AI tools offer superior benefits compared to traditional learning resources?}
The promise of GenAI rests on the claim that it enables faster, more effective task performance than conventional resources. Yet rigorous evidence remains scarce. By comparing outcomes of knowledge workers assigned to LLM-assisted self-study with those restricted to traditional training resources, we test whether GenAI delivers superior performance outcomes. Beyond performance scores, we also assess engagement, dropout patterns, and resource preferences to capture knowledge workers’ revealed judgments about the relative value of LLMs.

\textbf{RQ3: Do generative AI tools function as inequality-reducing or inequality-expanding mechanisms in the acquisition of task-specific expertise?}
Prior work offers conflicting predictions. Some argue that weaker knowledge workers will benefit disproportionately, compressing performance distributions; others warn that already advantaged knowledge workers will leverage GenAI more effectively, widening gaps. We therefore test whether access to LLMs narrows or widens disparities, and along which dimensions of “ability” inequality manifests. In particular, we distinguish between conventional background markers and emergent AI-specific competencies.

\textbf{RQ4: What mechanisms underlie the ways in which generative AI tools exacerbate or reduce inequality in task performance?}
Even if GenAI shifts average outcomes, understanding how it does so is essential for managerial decision-making and operations design. We examine the mechanisms driving distributional effects, focusing on four in particular: mismatches between knowledge workers’ preferences and their actual abilities, scaffolding interventions such as conceptual maps, peer-learning, and the consequences of additional time-on-task. By probing these mechanisms, we identify not only whether GenAI alters inequality but also the channels through which it does so.

\section{Experimental Overview}


We design a randomized adoption study that mirrors how organizations evaluate GenAI for knowledge-work training and task execution under time and budget constraints. The study proceeds in four stages, mapping to a typical enterprise rollout: profiling, baseline assessment, a structured upskilling intervention, and post-intervention evaluation.

\textbf{Profiling and context.} Participants are profiled on role-relevant background characteristics (e.g., tenure proxies, prior domain exposure, self-assessed readiness, and learning preferences). This establishes covariates for heterogeneity analyses and enables calibration checks between perceived and actual capability.

\textbf{Baseline assessment.} A brief, task-relevant assessment provides an objective pre-intervention baseline. Items span (1) awareness and terminology to prevent floor effects and (2) conceptual/technical content that differentiates novice from advanced performers. Scores are normalized to enable comparison with the post-intervention assessment.

\textbf{Upskilling intervention (randomized).} Over a short, fixed horizon (e.g., three consecutive workdays with a minimum daily time budget), participants complete a self-directed upskilling program aimed at rapidly acquiring and applying new technical knowledge for a concrete deliverable. We randomize access to:

(*) A GenAI condition (LLM-assisted self-study and problem solving), and

(*) A baseline condition (traditional resources such as manuals, articles, recorded trainings, and peer discussions; no LLM access).

To examine implementation levers available to managers, we layer simple process variants within the GenAI condition for less-experienced participants: a scaffolding arm (conceptual roadmaps and recommended sequencing), a time-on-task arm (higher hourly budget), and a peer-collaboration arm (structured collaboration guidelines). These variants represent common organizational choices—workflow templates, time allocation, and teaming—that could shape realized value from GenAI.

\textbf{Post-intervention evaluation.} A second assessment captures task performance after the intervention, using overlapping but not identical content to measure learning and application rather than recall. Scores are normalized to the unit interval. In addition to performance, we record engagement and completion (dropout) as revealed-preference indicators of perceived resource value. All treatment effects are estimated with controls for baseline performance and pre-specified covariates; heterogeneity is analyzed with interaction terms that reflect plausible moderators in organizational settings (e.g., prior knowledge, GenAI-specific interaction capability, and process variant).

\textbf{Outcomes and decision mapping.} The primary outcome is individual task performance (a proxy for knowledge-work productivity). Secondary outcomes include engagement/attrition and resource preference. This bundle aligns with managerial questions about average impact, dispersion (equity/consistency), and adoption frictions, and it supports downstream cost–benefit or ROI translation at the team level.

\textbf{AI Interaction Competence} In this work, AIC  is defined as the set of practical skills and habits that enable an individual to use generative AI systems effectively and reliably in their work. It includes the ability to:

(1) Formulate precise and goal-oriented prompts that communicate intent clearly to AI systems.

(2) Filter and verify generative AI's outputs for accuracy, bias, and relevance rather than accepting them at face value.

(3) Iterate interactively, refining inputs and combining human judgment with machine suggestions to improve quality and efficiency.


In practice, AIC determines how well employees can translate GenAI access into measurable performance gains.

\section{Methodology}

Our experimental setting approximates the conditions faced by early-career knowledge workers adapting to GenAI tools. Although the participants were university students, the task environment—rapidly acquiring and applying unfamiliar technical information under time constraints—closely mirrors cognitive work processes in professional domains such as consulting, analytics, engineering, and research. These activities rely on individual initiative, problem structuring, information synthesis, and interpretation of ambiguous outputs—the same foundational competencies that GenAI is designed to augment. Using students therefore allows us to isolate and measure the underlying mechanisms of AI-assisted knowledge work (interaction, verification, and self-directed learning) while maintaining tight experimental control and causal identification. In this sense, the study captures an early-stage adoption context that parallels how organizations train and evaluate human–AI collaboration capabilities among new or transitioning employees.

\subsection{Setting and Participants}
The intervention was conducted at Texas A\&M University with a total of 179 participants. Recruitment primarily targeted engineering majors, including Industrial and Systems Engineering, Computer Science, Materials Science and Engineering, Chemical Engineering, Mechanical Engineering, Biomedical Engineering, and Electrical and Computer Engineering. A small number of business students also participated after learning about the study informally, although they were not directly recruited.

The participant pool included undergraduates, master’s students, and PhD students, with the vast majority being undergraduates. Recruiting across multiple engineering disciplines was intentional: on the one hand, these students shared a broadly similar technical orientation; on the other hand, their disciplinary diversity allowed us to distinguish more clearly between novice and advanced learners for the purposes of analysis.

\subsection{Motivation for the self-study topic}
The choice of large language models (LLMs) as the self-study domain was intentional and strategic. Several considerations motivated this decision:

1. Layered knowledge structure. LLMs lend themselves naturally to a two-layer distinction: a general ML layer (covering concepts like supervised vs. unsupervised learning, regularization, evaluation metrics) and a specific layer (covering the architectures, training regimes, and inference properties of LLMs). This separation allowed us to disentangle how general background knowledge and topic-specific expertise interact with GenAI use.

2. Architectural breadth. LLMs encompass a diverse set of subtopics --- such as attention mechanisms, reinforcement learning from human feedback, and self-supervised pretraining --- each of which is sufficiently rich to stand as a unit of study. This complexity made the domain especially well-suited for testing scaffolding interventions, including conceptual maps designed to help novices organize and sequence their study.

3. Participant motivation. Voluntary self-study requires strong intrinsic motivation. Survey responses and anecdotal evidence indicate that participants viewed LLM knowledge as not only professionally indispensable but also personally valuable and timely—something they were eager to learn regardless of study participation. Choosing LLMs as the study topic ensured sustained engagement and minimized attrition relative to a less relevant or abstract domain.

Finally, the recursive nature of the design --- participants using LLMs to study LLMs --- created a unique opportunity to examine how a technology mediates its own uptake, legitimacy, and effectiveness as a learning resource.

\subsection{Pre-Intervention Survey}
Prior to randomization, participants completed a structured survey capturing demographic and background information. Key variables included:

\begin{itemize}
    \item Age
    \item Gender
    \item GPA
    \item Learning preferences
    \item Self-assessed knowledge of machine learning (ML)
    \item Self-assessed knowledge of LLM Interaction Competence (AIC)
    \item Self-assessed knowledge of LLM internal architectures
    \item Future academic and career goals
\end{itemize}

The tripartite decomposition of LLM knowledge into (1) general ML knowledge, (2) Interaction Competence (AIC), and (3) architectural understanding was deliberate. It enabled us to conduct a layered analysis disentangling the effects of domain-general versus topic-specific knowledge on subsequent performance outcomes.

\subsection{Pre-Intervention Exam}
Participants next completed a 15-item multiple-choice exam designed to establish a baseline of technical knowledge. The exam was constructed along two orthogonal design dimensions. 

First, a \textbf{familiarity-anchor dimension}: to avoid floor effects and ensure that nearly all participants could answer at least some questions, the exam included 2--4 awareness items that anyone who had previously interacted with an LLM would likely know (e.g., what ``hallucinations'' are or common shortcomings of LLMs). These questions served both to sustain engagement and to provide a baseline measure of familiarity distinct from technical competence. 

Second, a \textbf{generality--specificity dimension}: the remaining items were split between (a) general machine-learning concepts (e.g., supervised vs.\ unsupervised learning, overfitting/regularization, evaluation notions) and (b) LLM-specific mechanics (e.g., architectural components, training pipeline, inference behaviors, prompting characteristics). This separation allowed us to disentangle the contributions of domain-general ML knowledge versus topic-specific LLM knowledge to subsequent learning outcomes. 

Participants were allotted 30 minutes to complete the exam and were instructed not to use external resources. Importantly, they were not told how the pre-intervention exam would be used in the study design, in order to minimize strategic behavior. Participants expressing concern about their lack of ML/LLM background were reassured that no prior expertise was expected.

\subsection{Intervention and Randomization}
Based on pre-intervention exam performance, participants were classified into two categories:
\begin{itemize}
    \item \textbf{Advanced learners} -- participants answering 8 or more items correctly.
    \item \textbf{Novice learners} -- participants answering fewer than 8 items correctly.
\end{itemize}

All participants were assigned the same self-study task: to learn as much as possible about LLMs, including their architectures, strengths and limitations, and the underlying ML techniques. Participants were instructed to dedicate a minimum of 3 hours per day for 3 consecutive days to self-study. 

Randomization was conducted \textit{in situ} (on-site, at the time of the experiment) in two stages:
\begin{enumerate}
    \item \textbf{Primary assignment} -- Participants were randomly allocated to either the Baseline condition (non-LLM resources) or the LLM condition (restricted to using LLMs).
    \item \textbf{Secondary assignment} -- Within the LLM condition, novice participants were further randomized into one of four scaffolding sub-conditions.
\end{enumerate}

The resulting treatment arms were as follows:
\begin{itemize}
    \item \textbf{Group A (Baseline)} -- Allowed to use any resources except LLMs (e.g., textbooks, YouTube, peers, online articles). Subdivided into A\textsubscript{novice} and A\textsubscript{advanced}.
    \item \textbf{Group B (LLM)} -- Restricted to using only the free version of ChatGPT (paid subscriptions prohibited). Subdivided into B\textsubscript{novice} and B\textsubscript{advanced}.
\end{itemize}

Within B\textsubscript{novice}, four scaffolding interventions were introduced:
\begin{itemize}
    \item \textbf{B\textsubscript{novice,baseline}} -- Same instructions as B\textsubscript{advanced}.
    \item \textbf{B\textsubscript{novice,time}} -- Required to study 4 hours per day instead of 3.
    \item \textbf{B\textsubscript{novice,scaffolding}} -- Provided with a high-level conceptual roadmap of essential topics (see Appendix), including recommended sequencing.
    \item \textbf{B\textsubscript{novice,peer}} -- Permitted to study collaboratively with other subgroup members; required to document with whom and how often they studied.
\end{itemize}

To reduce attrition risk, especially among novice participants, all participants were permitted to bring a one-page ``cheat sheet'' to the post-intervention exam. This was designed to emphasize comprehension rather than rote memorization.

\subsection{Post-Intervention Exam}
At the conclusion of the self-study period, participants completed a post-intervention exam consisting of 28 multiple-choice questions focused exclusively on LLM knowledge; that is, general ML questions were not included in the post-intervention exam. Several items overlapped with the pre-intervention exam to enable measurement of knowledge gains. Participants were allotted one hour to complete the exam; all finished within the time limit.

To prevent ties in exam scores (important for bonus allocation), three additional open-ended numerical items were included. These questions were deliberately non-trivial and designed to be unsolvable with precision, ensuring variability in answers. In tie situations, the participant whose numerical responses were closer to the true values received the higher ranking.

\subsection{Compensation and Incentives}
All participants received a base payment of \$50 upon completing the study. To incentivize genuine engagement during the self-study period, we additionally offered performance-based bonuses: the top 10\% of participants on the post-intervention exam received an extra \$150. This incentive structure was designed to encourage sustained effort and maximize internal validity of the intervention.

\subsection{Outcome variable.}  
Because the pre-intervention and post-intervention exams contained different numbers of questions, raw scores are not directly comparable. We therefore normalized performance by computing the fraction of available points earned:
\[
Y_{i} = \frac{\text{points scored}_{i}}{\text{questions available}_{i}}.
\]
For the pre-intervention exam, each item contributed one point if answered correctly. For the post-intervention exam, participants could select multiple answers, with scoring rules designed to discourage random guessing. Each question yielded 1 point if fully correct, 0.5 points if partially correct (some but not all correct options selected), and 0 if incorrect. This ensured that both exams are scaled to the unit interval and directly comparable, while rewarding partial understanding and penalizing indiscriminate answering.

 \section{Results}

 \subsection{RQ1: Do knowledge workers’ perceptions of their own knowledge align with their actual knowledge levels?}

The first step in assessing how participants engaged with the study was to compare their self-assessed knowledge to their measured performance. Across the sample, we found encouraging evidence of consistency: participants’ self-assessments were not random guesses or aspirational self-presentation, but correlated meaningfully with their actual scores. \textbf{This serves as an important form of internal validation.} It suggests that participants responded in good faith, providing reliable inputs rather than inflating or deflating their reported competence.

The strength of this alignment, however, varied by domain. For self-assessments of general machine learning knowledge, the correlation with test performance was the strongest ($\rho = .71$, $p < .01$). Participants appeared able to accurately evaluate their mastery of broad ML concepts such as supervised learning, overfitting, and evaluation. For LLM-specific knowledge, the correlation remained positive and significant but weaker ($\rho = .6$, $p < .01$), indicating that participants could still gauge their grasp of architectural and functional details, though with more noise, potentially due to mild lack of understanding what LLM architectural knowledge truly entails. By contrast, AIC showed only a modest association with measured performance ($\rho = .46$, $p < .05$). Importantly, AIC was not itself a domain of self-assessment but rather inferred from behavioral performance, making this weaker correlation a reflection of skill measurement rather than miscalibration.

In other words, participants knew themselves well on the dimensions they could judge --- but the ultimate relevance of this calibration was limited. While general ML and LLM knowledge were positively correlated with actual baseline performance, these forms of prior knowledge did not moderate treatment gains: novices and advanced learners improved at similar rates once baseline knowledge was accounted for. What mattered instead was AIC --- the practical skill in prompting and filtering that participants were neither asked to self-assess nor tested on in the pre-intervention exam. This creates a telling asymmetry. \textbf{The dimensions on which participants could introspect accurately were not the ones that drove inequality in outcomes while the skill that decisively shaped heterogeneity remained invisible to conventional exams.} 
 The implication is clear: self-assessments and prior knowledge exams can validate internal consistency, but they are insufficient for anticipating who benefits most from AI-mediated learning.


\subsection{RQ2. To what degree do generative AI tools offer superior benefits compared to traditional learning resources?}

The central comparison in this study is between participants assigned to LLM-assisted self-study and those restricted to traditional resources. Across multiple outcomes, the results consistently demonstrate that access to generative AI tools conferred substantial advantages.

\textbf{Performance gains.} Participants in the LLM condition achieved significantly higher scores on the post-intervention exam than their peers in the baseline group. The average post-intervention exam score among LLM participants was $M = .56, SD = .26$, compared to $M = .48, SD = .23$ in the baseline group. To account for baseline differences, we estimated an OLS regression using $A_{\text{baseline}}, B_{\text{baseline}}$ (treatment = 0) and $A_{\text{LLM}}, B_{\text{LLM}}$ (treatment = 1), with controls for normalized pre-intervention exam score, GPA, gender, and self-reported study preferences. \textbf{The treatment coefficient remained positive and statistically significant ($p < 0.05$). These results indicate that the benefits of LLMs were not an artifact of pre-existing differences between groups, but robust across specifications.}

\textbf{Attrition as an outcome.} Beyond test scores, the most striking asymmetry between groups emerged in patterns of dropout. Overall, 20 participants in the baseline group withdrew before the post-intervention exam, compared to only 9 in the LLM group. A chi-square test showed that assignment to the baseline condition was a strong predictor of attrition ($p < 0.01$). Baseline participants most commonly cited illness or scheduling conflicts, but the systematic imbalance suggests otherwise: the absence of LLM access undermined motivation to persist. In effect, attrition became a form of revealed preference, signaling that participants judged the baseline condition as less worthwhile. This pattern not only biased final outcomes in favor of the LLM group but also illustrates how technology availability can shape engagement itself.

\textbf{Participants' preferences.} Self-reported preferences mirrored these behavioral patterns. When asked which resources they preferred for learning, 69\% of participants selected LLMs, surpassing lectures (64\%), YouTube tutorials (67\%), and textbooks (40\%). This finding underscores that generative AI tools are not merely a novel option but have rapidly become the most desired study resource among participants. Preferences and attrition reinforce each other: participants not only say they prefer LLMs but also act on those preferences when denied access, by disengaging from the study altogether.

\textbf{Validity checks.} All analyses use the completer sample (post-intervention exam scores available), so estimates are not mechanically driven by missing outcomes. Randomization balance was assessed with Mann--Whitney U tests comparing $A_{\text{baseline}}, B_{\text{baseline}}$ to $A_{\text{LLM}}, B_{\text{LLM}}$; we found no statistically significant differences in demographic or educational attributes. Differences between novice and advanced learners are accounted for by controlling for the pre-intervention score  in all specifications, and heterogeneity by baseline knowledge is examined directly via the Treatment $\times$ Pre-intervention exam score interaction analysis reported in RQ3.

Taken together, the results for RQ2 demonstrate how Generative AI tools substantially outperformed traditional resources, not only in terms of exam performance but also in maintaining participant engagement. Participants in the baseline condition were disadvantaged on every dimension: they learned less, preferred their resources less, and dropped out at higher rates. These findings highlight a critical shift: LLMs are not just incrementally better --- they are already perceived and experienced by participants as indispensable for effective self-study.

\subsection{RQ3. Do generative AI tools function as inequality-reducing or inequality-expanding mechanisms in the acquisition of task-specific expertise?}

A central debate in the literature is whether generative AI reduces or exacerbates inequality in cognitive performance. On the one hand, prior studies suggest that weaker performers benefit disproportionately, leading to compressed outcome distributions. On the other, evidence has emerged that GenAI may amplify inequality by rewarding those already advantaged. Our findings provide a nuanced perspective: generative AI did not systematically reinforce traditional academic hierarchies, but it did generate new fault lines in performance.

\textbf{Traditional academic markers: GPA and prior knowledge.}
We first tested whether GPA moderated the treatment effect of LLM use. The interaction between GPA and treatment assignment was small and statistically insignificant ($p=0.59$). High-GPA participants in the LLM condition improved, but no more than their lower-GPA peers. Similarly, pre-intervention exam scores, which measure prior topic-specific knowledge, did not significantly interact with treatment status ($p=0.2$). In other words, LLMs did not preferentially reward those who already possessed stronger academic backgrounds or prior familiarity with the material.

This null result is itself notable. It challenges the widespread assumption that technological interventions magnify existing prior-knowledge gaps by disproportionately benefiting high achievers. Instead, GenAI appears to flatten these traditional dimensions of inequality: both strong and weak participants gained from access, and the gap between them did not widen.

\textbf{The emergence of a new inequality axis: AI Interaction Competence.}
While GPA and prior knowledge did not matter, a very different picture emerged when we examined AIC. Participants who demonstrated high AIC at baseline --- i.e., those adept at formulating effective prompts, filtering responses, and applying outputs --- benefited disproportionately from the treatment. In interaction analysis, the coefficient on the (Treatment $\times$ AIC) term was positive and significant ($p < 0.05$).

\textbf{Unlike GPA or pre-intervention exam score, which showed no meaningful interaction with treatment, AIC emerged as the decisive moderator of outcomes. AIC thus supplanted traditional academic credentials as the most consequential moderator of performance.} In other words, what mattered most was not years of coursework, grades, or accumulated disciplinary knowledge, but rather the fluency with which participants could harness a generative AI system.
This marks a profound shift. For decades, educational inequality has been framed through familiar dimensions --- GPA, prior subject mastery, access to tutoring, socioeconomic background. Our findings suggest that a new dimension is rapidly becoming decisive: AI literacy as embodied in AIC. This is a skill not formally taught, not evenly distributed, and not easily signaled by existing academic markers. Yet in the presence of generative AI, it determines who advances and who falls behind. In this sense, AIC is not merely another predictor --- it is the emergent “currency” of learning in AI-mediated environments.

\textbf{Conditional effects: the three-way interaction.}
Having established AIC as the dominant moderator, we next asked whether it could account for the otherwise absent heterogeneity between novice and advanced learners. To investigate this, we conducted a three-way interaction analysis including Treatment , pre-intervention exam score, and AIC. The interaction term was statistically significant ($p < 0.05$). For ease of interpretation, we rescaled the pre-intervention exam score which yielded a positive coefficient: \textbf{novice learners (i.e., those with low prior knowledge) benefited more from LLM-assisted study, but only if they had high AIC levels}. In contrast, novices with low AIC levels showed substantially smaller gains, despite having more “room to grow."

\textbf{Distributional consequences.}
Taken together, these results indicate that generative AI compressed inequality along conventional academic lines (GPA, prior knowledge) but expanded it along a novel dimension: the ability to interact effectively with AI. Moreover, the three-way interaction highlights that the benefits of LLMs for weaker participants are conditional: novices can surpass ahead if they are strong extractors, but remain disadvantaged if they lack this skill. This is inequality of a new kind. It is not about who has studied more or who has higher grades, but about who possesses an emerging literacy that is neither systematically taught nor evenly distributed.
\begin{figure}[tb]
\centering
\includegraphics[height=2.6in, width=3.9in]{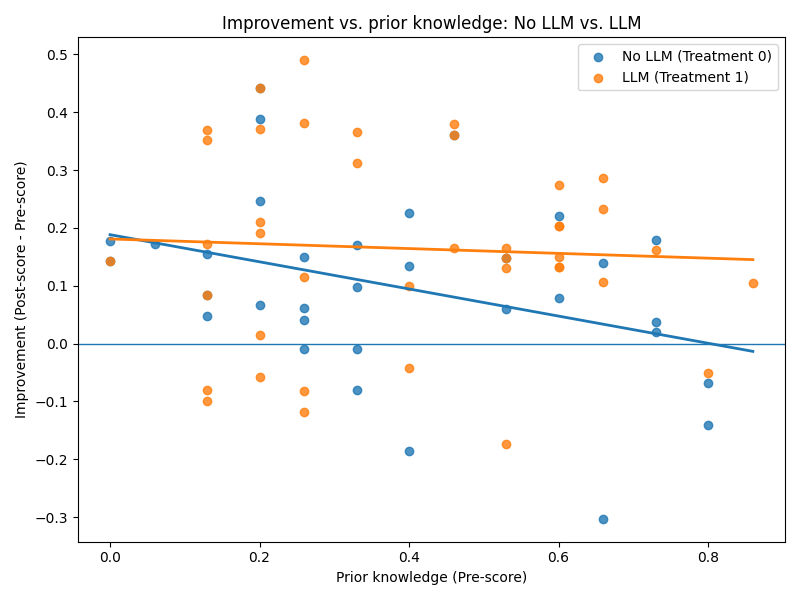}
\caption{ Learning Gains by Baseline Knowledge: No-LLM vs. Standard LLM.}
\label{fig:fig1}
\end{figure}

\textbf{Learning gain pattern} Another interesting insight on the relationship between baseline knowledge and learning gains is that this relationship appears to differ across conditions. Although the formal interaction between treatment and pre-intervention exam score is not statistically significant in our main specification, exploratory analyses suggest that the relationship between baseline knowledge and learning gains may differ across conditions. In the no-LLM condition, gains decline with prior knowledge: participants who start with lower pre-intervention scores improve the most, while those who begin with higher prior knowledge show smaller gains. This is consistent with a standard room-to-improve or diminishing-returns pattern. Under conventional self-study, lower-baseline participants have many obvious gaps to close, whereas higher-baseline participants may reach the limits of what fixed resources can add more quickly. By contrast, in the standard LLM condition, this downward relationship is weaker. Gains appear more evenly distributed across the prior-knowledge range, suggesting that LLM access may attenuate the usual tendency for learning gains to taper as baseline knowledge rises.

One possible implication is that LLMs may change not only the level of learning, but also its shape. In a traditional-resource environment, learning is relatively bounded: participants work through a finite set of materials, and the marginal value of additional study may decline quickly once the most important concepts have already been mastered. In an LLM-assisted environment, however, the learning process becomes more generative and expandable. Participants can request tailored explanations, ask follow-up questions, compare concepts, generate examples, test partial understanding, and continuously redirect the material toward their own confusions. In that sense, the knowledge space is less bounded. Rather than simply reviewing a fixed body of content, participants may be able to keep extending and reorganizing their understanding dynamically. This could help explain why the no-LLM group shows a clear diminishing-returns pattern while the LLM group appears flatter.

At the same time, the data also suggest that this expanded learning space is not equally usable for everyone. Participants with very low prior knowledge do not seem to gain an obvious additional advantage from standard LLM access relative to control. A plausible mechanism is that open-ended LLM use creates navigational demands of its own. To benefit from the tool, users may need enough conceptual grounding to know what to ask, how to interpret answers, how to recognize when an explanation is incomplete or misleading, and how to convert generated output into stable understanding. Participants with moderate or higher prior knowledge may be better able to exploit the flexibility of the LLM environment, whereas those with very low prior knowledge may struggle to turn that flexibility into effective learning. This interpretation also fits the broader argument in this work that access alone is not enough and that AI-specific interaction skill matters for realized gains.


\subsection{RQ4. What mechanisms underlie the ways in which generative AI tools exacerbate or reduce inequality in task performance?}

The aggregate treatment effects mask important heterogeneity in the processes through which inequality was either amplified or reduced. Here we focus on three mechanisms documented in the data: (1) mismatches between participants’ preferences and their actual ability to use LLMs effectively, (2) the impact of scaffolding via conceptual maps, and (3) the consequences of extending mandated study time.

\textbf{Preference–ability mismatches.}
One of the clearest findings in this study is that preference does not equal ability. To examine this relationship, we compared AIC levels between participants who reported preferring LLMs as their study method and those who preferred other methods. A Mann–Whitney U test revealed a statistically significant difference ($p < .05$), with a moderate effect size: participants who preferred LLMs generally exhibited higher AIC levels. However, the correlation between preference and ability was only small to moderate ($r=0.26$), suggesting that the alignment between preference and actual competence was incomplete.

This imperfect overlap gave rise to two distinct subgroups.

Group 1: Enthusiastic but unskilled. Participants with relatively low AIC levels nonetheless expressed a preference for LLMs. The behavior of this group is less well understood and warrants further study, as it is unclear whether their preference reflects optimism, overconfidence, or a lack of awareness of their limitations.

Group 2: Skilled but reluctant. Participants with high AIC levels reported preferring other methods of study. Closer inspection of this group revealed an interesting pattern: these participants frequently attended lectures (as self-reported in the survey) even when attendance was optional and often reported a preference for peer-based learning. This suggests that their reluctance to rely on LLMs may stem not from inability but from a desire for the social and collaborative dimensions of learning --- something current LLMs cannot provide.

To probe this hypothesis, we estimated a logistic regression model including social-learning preference as a predictor and LLM preference as the outcome. Although the coefficient was positive, consistent with the hypothesis, it did not reach conventional levels of significance ($p \approx 0.1$). The small number of participants in Group 2 likely limited statistical power.

Taken together, these results highlight that participant preferences cannot be taken at face value as proxies for competence. While there is some alignment --- LLM-preferring participants were, on average, more skilled --- a nontrivial share of participants either overestimated their ability (Group 1) or chose against their demonstrated strengths (Group 2). These mismatches complicate the integration of GenAI into knowledge-work training: they suggest that access alone will not guarantee equity, since participants’ decisions about how to learn do not always align with where they learn best.

\textbf{The scaffolding intervention: conceptual maps as equalizers.}
The most striking evidence of inequality reduction came from the scaffolding intervention. Novice participants provided with a conceptual roadmap of LLM topics performed modestly but significantly better than unguided novices. The average post-intervention exam score among scaffolded novices was $M = .45, SD = .14$, compared to $M = .38, SD = .22$ in the baseline LLM group, with a weakly significant difference ($p > .05$, $p < .10$). The consistency of the effect suggests that conceptual scaffolding may play an important complementary role in hybrid self-study settings.

The more revealing result, however, emerged when we tested for interaction effects. An OLS regression comparing scaffolded novices to baseline LLM novices included an interaction between treatment type (0 = baseline LLM, 1 = conceptual scaffolding) and AIC. The coefficient on the interaction term was negative and weakly significant ($p > .05$, $p < .10$), indicating that the marginal benefit of scaffolding diminished as AIC increased. \textbf{Put differently, conceptual scaffolding had its greatest impact on participants with low AIC, while high-skill participants derived little additional benefit.}

\textbf{This pattern points toward a critical mechanism: scaffolding compresses heterogeneity in outcomes by disproportionately aiding weaker extractors.} Whereas unguided LLM use amplified differences in AIC levels, the roadmap reduced these disparities by providing a structured path through the material. Variance in post-intervention exam scores was lower among scaffolded novices ($SD=.14$) compared to unguided novices ($SD=.22$), consistent with the view that scaffolding reduces inequality by leveling the floor.

\textbf{The additional time intervention: diminishing returns.}
By contrast, extending the daily study requirement from three to four hours produced no significant performance gains. Novices in the “more time” group scored $M = .37, SD =.22$, statistically indistinguishable from unguided LLM novices, $M= .38, SD= .23$. Ordinary least squares (OLS) models comparing the 
$B_{novice,baseline}$,
$B_{novice,time}$
 groups --- with controls for demographics, pre-intervention score, and Interaction Competence --- found no significant treatment coefficient for the additional time condition.

This null result is revealing. It suggests that simply increasing the quantity of exposure does not reduce inequality when participants lack efficient study strategies. Without scaffolding or high AIC levels, more time-on-task produced diminishing returns, and in some cases, participants reported frustration with the extended requirement.

 \section{Discussion}

 Our findings offer a new perspective on how generative AI reshapes the production function of knowledge work. At a fundamental level, GenAI alters what it means to be “able.” Traditional indicators of capability—prior knowledge, GPA, or accumulated expertise—lose much of their predictive power once access to large language models is universal. Instead, performance hinges on a different skill set: the ability to elicit, interpret, and operationalize machine-generated outputs. As defined earlier, AI Interaction Competence captures this interactional fluency and determines whether technology amplifies or attenuates individual performance. The emergence of AIC explains why GenAI can simultaneously equalize and stratify outcomes: it lowers barriers to entry by democratizing access to expertise, yet rewards those who can communicate effectively with algorithms, generating a new fault line in organizational capability.

Viewed through the lens of human–AI complementarity, these dynamics extend established theories of technological change. Prior work in information systems and economics shows that general-purpose technologies raise average productivity but initially increase variance because returns depend on complementary human and organizational assets. GenAI exemplifies this pattern. The technology itself is accessible and inexpensive, but its benefits depend on interactional fluency—a capability that remains unevenly distributed. In this sense, GenAI is not merely a substitute for labor but a multiplier of competence: it magnifies existing differences in how individuals translate information into value. Organizations adopting AI therefore face a dual challenge: realizing average productivity gains while managing the variance those gains create.

From a managerial standpoint, our results point to a consistent pattern: GenAI access translated into an approximate 17\% productivity lift --- consistent with field evidence of AI-assisted performance gains in professional work settings. However, dispersion among novices increased by roughly 47\% in standard deviation—more than doubling the variance. Notably, a scaffolding intervention (conceptual roadmaps and sequencing) reduced this dispersion by nearly 40\% without lowering the mean, demonstrating that small process changes can stabilize outcomes. This variance is not random; it stems from uneven distribution of AIC levels across individuals. The central managerial challenge is therefore not access but alignment—ensuring that employees possess the interactional fluency required to convert AI outputs into consistent, high-quality performance.  Firms can address this gap through targeted micro-training that focuses on prompting logic, verification heuristics, and synthesis structure. Short interventions of 60–90 minutes can meaningfully raise the performance floor, especially when coupled with light scaffolds such as standardized prompt templates or review checklists. These process interventions—analogous to standard operating procedures (SOPs) in traditional operations—reduce performance variance by about one-third without diminishing the mean. The implication is clear: the effective management of GenAI hinges less on the technology itself and more on the design of complementary routines that embed consistency, discipline, and feedback into human–AI interaction.


These insights reframe AI adoption as a problem of capability design rather than tool procurement. Although our experiment used a controlled learning environment, the mechanisms we identify—problem framing, prompt design, verification, and synthesis—mirror the core activities of professional knowledge work such as analysis, writing, and decision support. The heterogeneity observed in our setting therefore parallels the dispersion managers encounter when deploying GenAI within organizations. Firms that systematically identify AI Interaction Competence gaps, embed short micro-trainings into onboarding and workflow cycles, and implement light scaffolds such as prompt templates or review checklists will capture GenAI’s benefits more consistently. Conversely, organizations that treat AI primarily as a cost-saving or automation tool risk widening internal capability gaps and undermining collective performance. Managing variance through training and process design is thus not a peripheral HR issue but a strategic capability: it aligns technological efficiency with organizational reliability and transforms GenAI adoption from a disruptive experiment into a repeatable, learning-driven source of advantage.

\bibliographystyle{plain}
\bibliography{genaibib2}
\end{document}